# Intelligent Magnetic Inspection Robot for Enhanced Structural Health Monitoring of Ferromagnetic Infrastructure

Angelina Tseng[1], Sean Kalaycioglu[2]
[1]Santa Catalina School, Monterey, California, USA
[2]Toronto Metropolitan University, Toronto, ON, Canada

*Abstract*—This paper presents an innovative solution to the growing problem of infrastructure deterioration in the United States, where approximately one-third of facilities are in poor condition, and over 130,000 steel bridges have exceeded their intended lifespan. Aging infrastructure, particularly steel structures, is vulnerable to corrosion and hidden defects, which pose significant safety risks. The Silver Bridge collapse, where an undetected flaw resulted in the tragic loss of 46 lives, underscores the limitations of traditional inspection methods reliant on manual human observation. These methods often fail to detect subtle or concealed defects, leading to catastrophic failures. In response to this urgent need for improved inspection technologies, this work introduces an AI-powered magnetic inspection robot. The robot is equipped with magnetic wheels that enable it to adhere to and navigate complex ferromagnetic surfaces, including vertical inclines, internal corners, and curved elements. Its robust mobility allows it to access hard-to-reach areas and conduct comprehensive inspections, offering a versatile solution for large-scale infrastructure assessment. Additionally, the system incorporates advanced machine learning techniques, utilizing MobileNetV2, a deep learning architecture, for defect detection. Trained on a large dataset of steel surface defects, the model achieved 85% precision, demonstrating strong performance across six distinct defect types. The robot's deep learning-driven inspection process significantly enhances the accuracy and reliability of structural assessments, surpassing traditional methods in terms of defect detection and operational efficiency. The findings of this project suggest that integrating robotic systems with AI-driven image analysis offers a transformative approach to infrastructure inspection, providing a scalable, automated solution that can reduce human labor, increase detection precision, and improve the safety and sustainability of critical infrastructure assets.

## I. INTRODUCTION

Aging civil infrastructure, particularly steel bridges, poses significant risks due to material degradation, including corrosion and fatigue, which may lead to catastrophic failures. A notorious example of such a failure is the collapse of the Silver Bridge in West Virginia in 1967, where an unnoticed fracture resulted in a fatal structural collapse. This event underscores the limitations of traditional visual inspections that rely heavily on human inspectors, who are constrained by physical access, fatigue, and subjectivity [1, 2]. With over 130,000 steel bridges across the United States requiring regular inspections, more advanced, objective, and comprehensive inspection methods have become crucial.

### A. Technological Advancements in Robotic Inspection Systems

Robotic inspection systems have emerged as an effective solution for addressing the limitations of manual inspection techniques. These systems utilize mobile robots equipped with cameras, sensors, and sophisticated software to provide real-time data analysis. Notably, robots can access areas of bridges that are difficult or dangerous for human inspectors to reach, including vertical and inverted surfaces [3].

Drones and magnetic adhesion robots have become increasingly popular for inspecting bridges. Drones are able to provide a wide coverage of aerial inspections, while magnetic adhesion robots offer the ability to inspect complex steel structures, adhering to surfaces using magnetic wheels or treads [4, 5, 6].

A significant advantage of robotic systems is their ability to operate continuously without suffering from fatigue, as is often the case with human inspectors. Additionally, the integration of advanced sensors, such as ground-penetrating radar (GPR) and infrared thermography, allows robots to detect both surface and subsurface defects, providing more accurate and objective assessments than traditional methods [7, 8]. For instance, Gibb et al. [9] demonstrated that sensor fusion, which combines data from multiple sources such as infrared cameras and ultrasonic sensors, improves the detection of structural flaws.

### B. Non-Destructive Evaluation (NDE) Techniques

Non-destructive evaluation (NDE) techniques have revolutionized the inspection process for aging infrastructure by providing ways to assess structural health without causing damage. A variety of NDE methods, including GPR, ultrasonic testing, and infrared thermography, have been successfully integrated into robotic platforms, allowing for comprehensive assessments of a structure's condition [10, 11]. NDE techniques have proven essential for identifying both surface and subsurface defects, such as cracks, voids, and corrosion [12].

Ahmed et al. [13] reviewed the state-of-the-art NDE technologies used in bridge inspections, focusing on the integration of these technologies into autonomous robotic systems. By utilizing multiple NDE methods simultaneously, robotic platforms can capture a wide range of data, improving both the accuracy and the reliability of inspections. For example, the fusion of visual, thermal, and acoustic data can help detect both visible and hidden flaws, making these robotic systems indispensable in infrastructure management [14, 15].

## C. Machine Learning and Artificial Intelligence (AI) in Structural Health Monitoring (SHM)

Machine learning (ML) and AI technologies have further enhanced the capabilities of robotic inspection systems. These systems use deep learning algorithms, particularly convolutional neural networks (CNNs), to automatically detect structural flaws from the visual and sensor data collected by robots. CNNs are particularly effective for pattern recognition, making them ideal for identifying cracks, corrosion, and other forms of material degradation [16, 17, 18].

Li et al. [19] analyzed the application of deep learning algorithms in SHM, finding that CNNs are the most commonly used models for detecting structural defects in steel bridges. The use of AI enables real-time analysis of data, reducing the time between inspection and maintenance. AI-driven SHM systems can also predict the future condition of structures, allowing for more effective maintenance planning and reducing the likelihood of unexpected failures [20, 21].

One of the most promising applications of AI in infrastructure monitoring is the development of digital twins. Digital twins are virtual replicas of physical structures that are continuously updated with real-time data from sensors. These models can simulate the structural behavior of a bridge under various conditions, providing valuable insights for preventive maintenance [22, 23]. Wang et al. [24] demonstrated the use of digital twins for large-scale infrastructure monitoring, highlighting their potential for improving predictive maintenance strategies.

## D. Magnetic Adhesion Robots and Autonomous Platforms

Magnetic adhesion robots have been at the forefront of robotic technology for infrastructure inspection. These robots can climb vertical or inverted steel structures using magnetic wheels or treads, providing access to areas that are difficult or impossible for human inspectors to reach. In addition to their mobility, these robots are equipped with NDE sensors, enabling them to detect both surface-level and subsurface defects [25, 26]. Hanamura et al. [27] optimized the design of magnetic wheels for these robots, improving their energy efficiency and allowing for longer operational times without compromising performance.

Besides magnetic adhesion robots, various other autonomous robotic platforms, such as drones and ground-based systems, have been developed to further enhance infrastructure inspection capabilities. Drones, in particular, are useful for inspecting large surface areas quickly, although they are limited by environmental factors such as wind and rain [28, 29]. Ground-based robots, including legged and climbing robots, offer versatile solutions for navigating uneven terrain and difficult-to-reach areas [30, 31].

While the use of autonomous robotic systems has increased dramatically in recent years, there are still challenges associated with their deployment in extreme environments. For example, drones may struggle to navigate in high-wind conditions, and ground-based robots may have difficulty traversing uneven or unstable surfaces [32, 33]. However, advancements in machine learning and AI, as well as the continued optimization of robotic hardware, are expected to mitigate these challenges and further improve the reliability and effectiveness of robotic inspection systems in the near future [34, 35].

The remainder of this paper is organized as follows:

*Section II* provides a detailed System Overview, outlining the key components and functionalities of the robotic inspection system used for structural assessment. This section presents the Structural Design Overview, focusing on the mechanical aspects and design considerations for the robotic platform. It also covers the Electronic Design Overview, which highlights the electronics, sensors, and communication systems integrated into the robotic system. *Section III* discusses the application of *Machine Learning Models*, describing the algorithms used for defect detection and data analysis. *Section IV* introduces *Mathematical Models*, detailing the theoretical framework and algorithms that underpin the operation and functionalities of the robotic system. This section will elaborate on the computational models used to predict, analyze, and optimize the performance of the inspection system. *Section V* presents *Experimental Results*, which validates the theoretical models discussed in Section IV. This section will include data and findings from field tests and simulations that demonstrate the effectiveness and accuracy of the robotic system in real-world scenarios. Finally, *Section VI* concludes the paper, summarizing the key findings and proposing directions for future research.

## II. SYSTEM OVERVIEW

The robotic system designed for magnetic inspection integrates sophisticated engineering principles that align with rigorous systems engineering methodologies. These methodologies underscore a systematic approach, structured through phases of Requirements Definition, Preliminary Design, Detailed Design, Integration, and Testing/Verification. The system is delineated into two primary components: Structural Design and Electrical & Electronics Design, each tailored to meet specific operational demands in challenging inspection environments.

### A. Structural Design Overview

*Chassis and Framework:* The core structural component of the AI-powered magnetic inspection robot is its chassis, designed to provide robustness while maintaining flexibility and lightness. The chassis is constructed from high-strength materials such as aluminum and composites known for their durability and lightweight properties, making them ideal for use in operational scenarios that require magnetic adhesion to vertical or complex surfaces. The modular nature of the chassis supports easy assembly and maintenance, essential for adapting to varied inspection environments (Figure 1).

*Magnetic Adhesion System*: A pivotal feature of the robot is its Magnetic Adhesion System, which utilizes high-strength neodymium ring magnets. These magnets are strategically embedded within the structure to optimize the balance between adhesive force and mobility, enabling the robot to cling securely to ferromagnetic surfaces during inspections. This system is critical for maintaining stability and operability on vertical or sloped surfaces, where traditional traction methods are ineffective (Figure 2).

Locomotion System: The robot's mobility is facilitated by a sophisticated Locomotion System that incorporates custom-designed wheels, which are an integral part of the magnetic adhesion system. These wheels are powered by high-torque servos, providing the necessary power to maneuver over diverse surface textures and geometries. This system ensures that the robot maintains adequate traction and stability, crucial for operational efficacy on uneven or inclined surfaces.

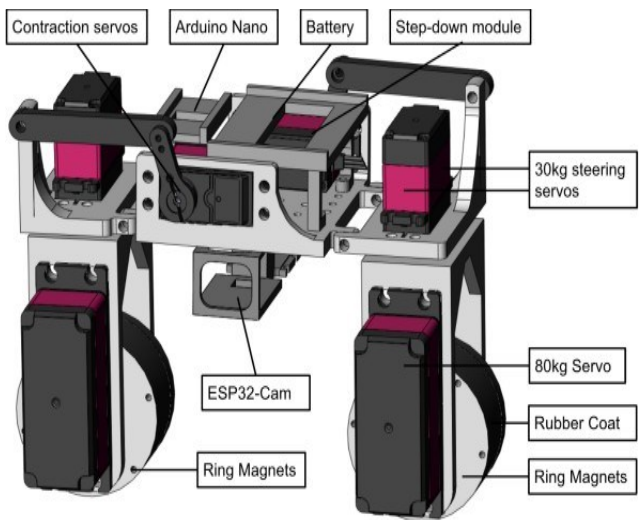
Fig. 1. The 3D model design of the robot

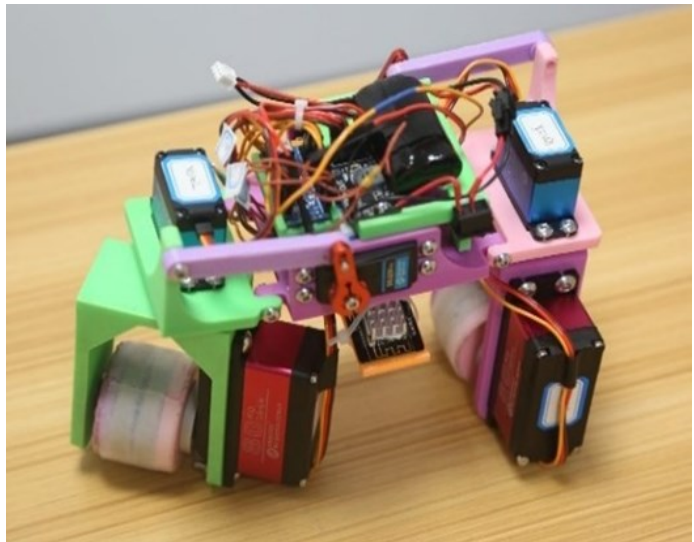
Fig. 2 Front view of the robot

### B. Electrical and Electronics Design Overview

*Core Components*: At the heart of the robot's electronics design are key components such as the ESP32-CAM Module and the Arduino Nano. The ESP32-CAM Module offers a compact, low-cost solution for image capturing and Wi-Fi communication, essential for real-time data transmission and remote operation. The Arduino Nano is employed for its versatility and ease of integration with various sensors and actuators, providing reliable control over the robot's movements and functionalities (Figure 3).

*Power System*: Energy efficiency and management are critical, addressed by integrating a high-capacity Li-Po battery that provides a lightweight yet robust power source. This is coupled with a Step-Down Voltage Regulator, which ensures that all components receive stable, optimized power by converting the battery's output to suitable voltages without significant losses, thereby enhancing the overall energy efficiency of the system.

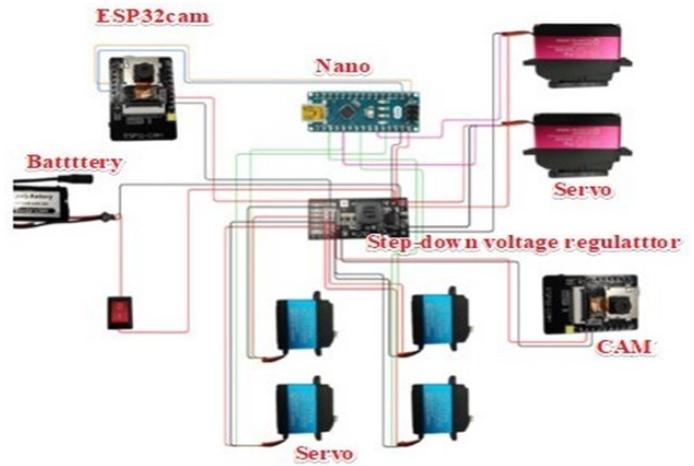
Fig. 3 Circuit Diagram

*Sensor Integration and Data Acquisition*: The robot is equipped with an array of sensors that facilitate advanced inspection capabilities. These include ultrasonic sensors for distance measurement, gyroscopes for orientation, and the integrated camera module of the ESP32-CAM for high-resolution imaging. This sensor array is critical for the robot's ability to detect defects, navigate autonomously, and execute its inspection tasks with high precision (Figure 4).

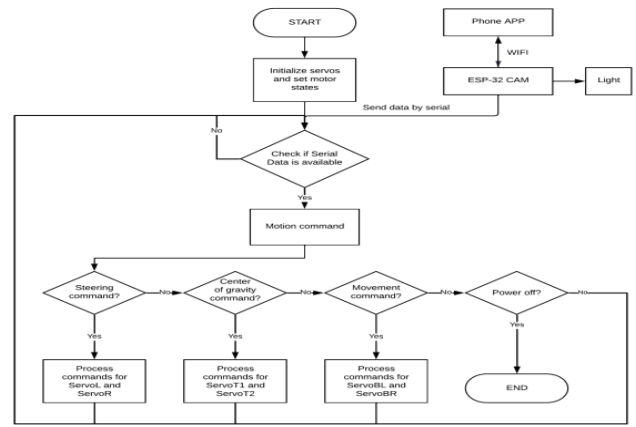
Fig. 4 Arduino Nano control logics

*Control Logic and Software Architecture*: The control logic, primarily managed by the Arduino Nano, is designed to handle complex command sequences and sensor integrations. This setup allows for precise control over the robot's steering mechanisms and adhesion systems, adapting dynamically to environmental inputs and inspection requirements. The accompanying software architecture is modular, allowing for flexibility in task execution and facilitating updates or modifications based on field requirements (Figure 5).

### C. Integration and Operation

*System Integration*: Integrating these components involves a systematic approach where each module is tested individually before being integrated into the larger system. This ensures that all parts function cohesively, adhering to the predefined requirements and operational standards.

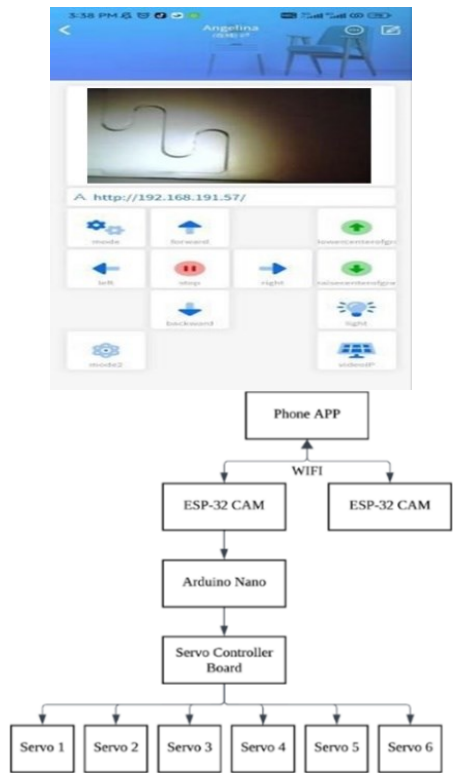

Fig. 5 Control system and Blinker interface

*Testing and Verification*: The final phase involves extensive testing and verification to ensure that the robot meets all operational requirements and is capable of performing under the expected conditions. This includes stress testing the magnetic adhesion on various ferromagnetic surfaces, validating the locomotion system's effectiveness on different geometries, and ensuring the reliability of the electronic systems under various operational scenarios.

Through these detailed and structured design and implementation phases, the AI-powered magnetic inspection robot is optimized to perform robustly and reliably, with minimal human intervention, in diverse and challenging inspection environments. This comprehensive system overview sets the stage for deeper exploration into the specific design details presented in the subsequent sections of this paper.

### III. MACHINE LEARNING MODELS

Advancements in machine learning have been pivotal in enhancing the precision and efficiency of defect detection systems in robotic inspections, particularly for steel structures. Among the various models, Convolutional Neural Networks (CNNs) and MobileNetV2 have been integral to the robot's defect classification and identification capabilities. Their deployment has significantly improved the reliability and accuracy of inspections. The dataset is a collection of high-resolution images taken from the surface of steel with several kinds of defects. Thousands of images are labelled and show typical defects that can appear under many forms, such as cracks, corrosion, or surface deformation, as shown in Figure 6.

#### A. Convolutional Neural Network (CNN)

*Model Architecture and Functionality*: The CNN employed in this project is designed with a multi-layer architecture that includes five convolutional layers followed by two fully connected layers. This design enables the CNN to learn hierarchical representations of spatial features from input images, which is crucial for recognizing patterns that indicate structural defects such as cracks, corrosion, or surface deformations [36]. The convolutional layers adaptively learn features from the images, while the fully connected layers aid in classifying these features into defect categories [Fig. 9].

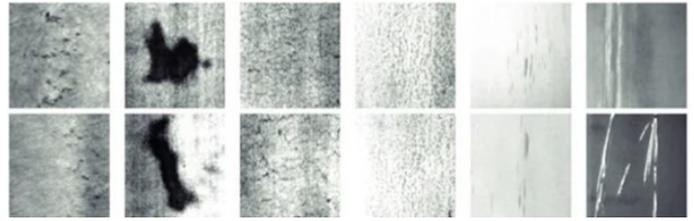

Fig. 6. Dataset pictures

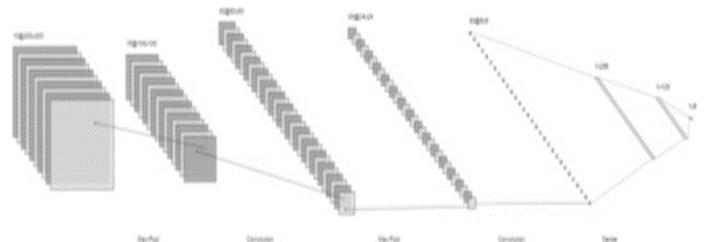

Fig. 7 CNN structure

*Training and Performance*: The CNN was rigorously trained on a dataset comprising 1,800 high-resolution images (200x200 pixels) labeled with various defect types. This dataset included diverse representations of common defects found on steel surfaces, ensuring comprehensive learning [Fig. 8]. Initially, the model achieved a training accuracy of 99%, indicating its ability to learn well from the training data. However, when deployed with real-world data from the robot's ESP32-CAM, the accuracy markedly decreased to about 14%, indicating challenges in generalizing the learned models to new, varied data sets.

*Challenges in Real-world Application*: The drop in performance with real-world data suggests the model's difficulty in handling variations in lighting, angles, and other environmental factors not fully represented in the training set. This highlights the need for more robust data augmentation and potentially retraining the model with more diverse and representative data from actual inspection environments.

#### B. MobileNetV2

Enhancements and Structural Design: In response to the limitations observed with the CNN, MobileNetV2 was integrated due to its suitability for mobile and embedded vision applications [37]. This model utilizes an inverted residual structure with linear bottlenecks that optimize both computational efficiency and accuracy, crucial for applications requiring low latency and high throughput [Fig. 8].

*Training Regimen and Adaptation*: MobileNetV2 was trained using the same dataset as the CNN to maintain consistency in learning scenarios. It underwent 50 training epochs with a training-to-validation ratio of 9:1, ensuring thorough learning and validation without overfitting. This model displayed a substantial improvement, achieving an accuracy of 85.71% on the field data, reflecting its robustness and adaptability to variable real-world conditions [Fig. 9].

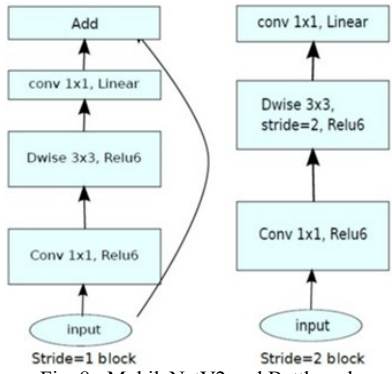

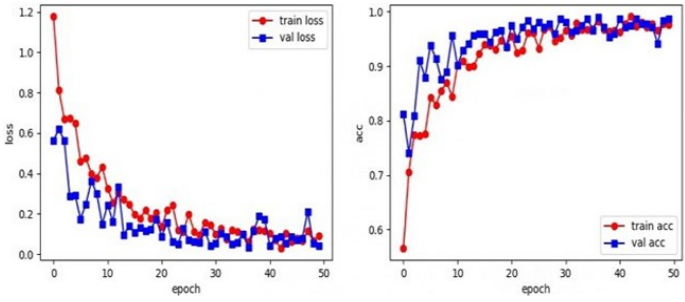

Fig. 8 MobileNetV2 and Bottleneck

Fig. 9 Loss and Accuracy Graphs

*Operational Efficacy and Real-world Reliability*: The enhanced accuracy of MobileNetV2 under real-world conditions confirms its efficacy and reliability for structural inspections. Its performance underscores the model's capability to handle variations in the dataset effectively, making it a superior choice for the dynamic and diverse conditions encountered during robotic inspections.

The integration of CNN and MobileNetV2 into the robotic inspection system underscores a significant advancement in automating defect detection in steel structures. This combination leverages the initial pattern recognition strength of CNNs with the added robustness and efficiency of MobileNetV2, addressing several real-world challenges effectively.

## IV. MATHEMATICAL MODELS

### A. Force and Torque Models

The design of the inspection robotic system necessitates sophisticated calculations to realize optimal performance, particularly concerning motor torque, magnetic adhesion, and dynamic forces. These calculations ensure the robot can efficiently navigate and adhere to various surfaces, including vertical and inverted orientations, under different operational conditions.

*1) Magnetic Force Calculations:*

The robot's ability to adhere to ferromagnetic surfaces, even when inverted, hinges on the strength of the magnetic forces generated by its magnetic wheels. The magnetic force (Fmagnetic) is calculated to ensure it can counteract the gravitational pull and any dynamic forces encountered during movement. The magnetic force is estimated using the following equation:

$$F_{magnetic} = \frac{B^2 \cdot A}{2 \cdot \mu_0} \quad (1)$$

where: B is the magnetic flux density (in Teslas), A is the area of contact area between the magnetic wheels and the surface in square meters, and μ0 is the permeability of free space (4π×10−7 H/m).

Considering the robot's total weight, the magnetic force must at least match the gravitational force exerted on the robot:

$$F_{gravity} = m \cdot g = 27.5 \times 9.81 = 269.78\,N \quad (2)$$

Thus, the magnetic force per wheel must be at least equal to this gravitational force, multiplied by a safety factor (chosen as 5 for engineering applications) to ensure stable adhesion, as in equation 3:

$$F_{magnetic\_total} = 5 \times 269.78 = 1348.9\,N \quad (3)$$

Given the robot employs two magnetic wheels, the required force per wheel is:

$$F_{magnetic\_per\_wheel} = \frac{1348.9}{2} = 674.45\,N \quad (4)$$

This force calculation ensures that the robot remains securely attached to the surface, even when operating upside down.

*2) Force Calculations for Motion*

The robot encounters various forces during movement, notably frictional and gravitational forces, especially on inclined or vertical surfaces. The frictional force (Ffriction) on a horizontal surface is defined as:

$$F_{friction} = \mu \cdot F_{normal} \quad (5)$$

where: μ is the coefficient of friction between the wheels and the surface, $F_{normal}$ is the normal force, which, for horizontal motion, equals the gravitational force. For vertical or inclined surfaces, the gravitational force component parallel to the plane is significant and is given by:

$$F_{gravity\_parallel} = m \cdot g \cdot \sin(\theta) \quad (6)$$

where θ is the angle of the inclination. Given the robot's total mass of 27.5 kg,

$$F_{gravity\_parallel\_vertical} = 27.5 \cdot 9.81 \cdot \sin(90°) = 269.78\,N \quad (7)$$

*1) Torque Calculations*

The torque τ required by the motors is critical for overcoming the combined frictional and gravitational forces, particularly on vertical surfaces. The torque depends on the wheel radius of the wheels r and the total force.

$$\tau = F_{total} \cdot r \quad (8)$$

Assuming a wheel radius of 0.1 meters and operating on a vertical surface where the dominant force is gravitational:

$$\tau = 269.78 \cdot 0.1 = 26.98 \, \text{Nm} \quad (9)$$

To ensure reliable operation under dynamic conditions (e.g., sudden movements or stops), a safety factor of 3 is applied, as in equation 10:

$$\tau_{safe} = 3 \times 26.98 = 80.94 \, \text{Nm} \quad (10)$$

Each motor must thus provide at least 80.94 Nm of torque to ensure the robot can navigate and adhere to surfaces under load without slipping. These detailed force and torque calculations are crucial in the design phase to guarantee that the robot performs effectively across all expected conditions.

*B. Extreme Locomotive Conditions*

The design and analysis of the robotic inspection system must consider critical factors of force and torque to ensure robust performance under extreme conditions. Such analyses focus on the mechanical limits and operational thresholds the robot can endure while maintaining functionality and stability, particularly in challenging environmental and operational scenarios.

*1) Force Analysis: Adhesive Force Calculation*

To ensure the robot can adhere to surfaces under minimal conditions, an initial calculation of the adhesive force between the magnetic wheels and the structure is critical. The robot's ability to sustain its own weight against a surface is contingent upon the balance between the adhesive force ($F_2$) and the robot's weight, factoring in the physical configuration of the wheels. The essential condition for maintaining adequate adhesive force is described by the following equation:

$$F_2 \times X_1 \times X_2 > P_h \Rightarrow F_2 > \frac{P_h}{X_1 \times X_2} \quad (11)$$

where $P_h$ represents the horizontal component of the gravitational force acting on the robot, and $X_1$ and $X_2$ are distances related to the robot's wheel configuration. This formula determines the minimum required adhesive force, $F_2$ that the robot must generate to counteract its weight and ensure stable adherence [Figure 10].

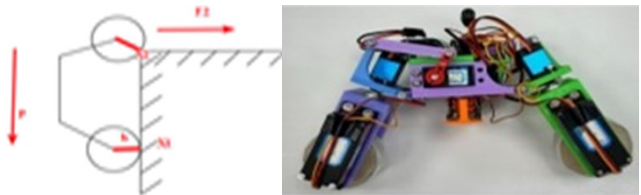

Fig 10. Extreme Force Calculation

Through practical analysis, it is demonstrated that the magnetic wheels must match or exceed an adhesive force of 2.227 N to guarantee stability. Empirical measurements indicate that the robot's adhesive capabilities substantially surpass this threshold, affirming its ability to maintain adhesion even under minimal force conditions. This ensures that the robot can securely operate on various surfaces without the risk of detachment.

*2) Torque Calculation for Critical Situations*

In scenarios where the robot encounters maximum load, such as traversing an internal corner between two orthogonal surfaces, calculating motor torque becomes essential. The robot must develop adequate torque to overcome the combined forces of adhesion and gravity acting on its barycenter. The minimum torque requirement for the front wheel to maintain stability is given by:

$$M_{moving} > r(F_{2.1} + kF_{2.2} + \frac{P}{2}) \quad (12)$$

where $M_{moving}$ is the moving motor torque, $r$ is the wheel radius, $F_{2.1}$ and $F_{2.2}$ are the forces acting on the front and rear wheels respectively, $k$ is a constant related to the robot's weight distribution, and $P$ represents the gravitational force. This equation underscores the need for the robot to generate sufficient torque to prevent slipping or detachment when maneuvering through complex geometries [Figure 11].

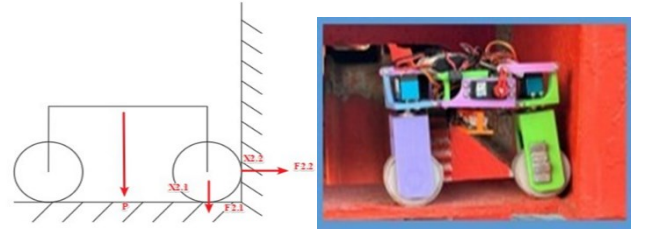

Fig. 11 Extreme Torque Calculation

The analysis dictates that the motors selected should be capable of providing at least twice the torque calculated theoretically, considering dynamic conditions for reliable operation. The motors chosen, 80 kg servos, are adequately robust, ensuring they can handle the robot's 27.5 kg load in all extreme locomotive scenarios.

By rigorously calculating the forces and torques necessary for operation under extreme conditions, the design ensures that the robot is not only capable of performing standard tasks but is also equipped to handle unexpected or high-stress environments, thereby enhancing its utility and reliability in practical field applications.

V. EXPERIMENTAL RESULTS

*A. Experiment 1: Maximum Load Capacity on Vertical Surfaces*

*Objective and Methodology*: The primary objective of this experiment was to assess the robot's load-bearing capacity and to establish the maximum load it could support while maintaining stable locomotion on a vertical steel surface. This assessment is crucial to determine the operational limits of the robot under increased mechanical stress, which mimics real-world conditions where the robot might need to carry additional equipment or sensors.

*Procedure and Setup*: The experiment involved incrementally increasing the load on the robot while it maneuvered on a standardized vertical steel surface. Each incremental load was maintained for a duration sufficient to assess stability, adhesion, and mobility. The robot's speed, adhesion stability, and any signs of mechanical strain or slippage were meticulously recorded.

*Results and Observations*: The results indicated that the robot maintained stable movement and adhesion with loads up to 25 kilograms. At this load, the robot operated efficiently without significant reduction in operational speed or adhesion quality. When the load was increased to 30 kilograms, a noticeable decrease in speed was observed, suggesting the upper limits of its operational capacity under current configurations.

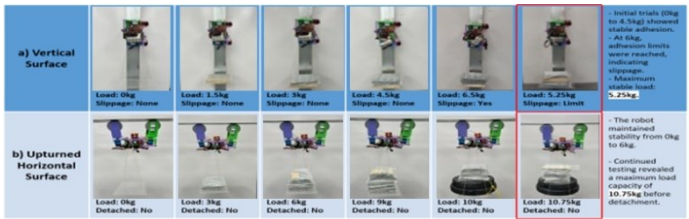

Fig 12. Maximum Load Capacity on Vertical Surfaces

*Conclusion*: From the experimental data, the maximum effective load for stable operation on vertical surfaces was determined to be 27.5 kilograms. This load represents a critical threshold, beyond which the robot's performance begins to degrade, particularly in terms of locomotion speed and stability. The findings are crucial for optimizing the robot's design and deployment in scenarios requiring variable load capacities. These results are graphically summarized in Figure 12, highlighting the relationship between load weight and locomotive efficiency on vertical surfaces.

*Significance*: Understanding the load-bearing capacity of the robot on vertical surfaces enables more accurate planning and utilization in industrial applications, ensuring that the robot can perform its designated tasks without risk of failure or inefficiency. This experiment lays the groundwork for subsequent modifications to enhance load capacity and overall stability, potentially expanding the robot's utility in more demanding industrial environments.

B. *Experiment 2: Surface Thickness and Adhesion*

*Objective and Methodology*: The objective of this experiment was to evaluate the robot's ability to adhere to and navigate ferromagnetic surfaces of varying thickness. Surface thickness plays a critical role in the strength of the magnetic force generated by the robot's magnetic wheels, directly affecting its adhesion and stability. Thicker surfaces generally allow for stronger magnetic attraction, while thinner surfaces reduce the robot's grip, potentially compromising performance in real-world applications where surface thicknesses may vary.

*Procedure and Setup*: The robot was tested on a range of ferromagnetic surfaces with varying thicknesses, including 3 mm, 5 mm, 7 mm, and 10 mm. For each thickness, the robot's movement, stability, and adhesion were observed and recorded. The experiment was conducted under consistent environmental conditions to ensure that any variations in performance were attributed to changes in surface thickness alone.

*Results and Observations*: The robot demonstrated optimal performance on surfaces with a thickness of 7 mm or greater. On these surfaces, the robot exhibited strong magnetic adhesion, stable movement, and full maneuverability. However, when tested on thinner surfaces, particularly those around 3 mm in thickness, the robot's adhesion decreased slightly. This reduction in magnetic force led to minor instability, particularly during sharp movements or changes in direction.

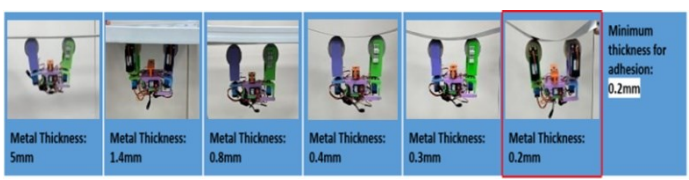

Fig 13. Surface Thickness and Adhesion

*Conclusion*: The results indicate that surface thickness is a crucial factor in the robot's ability to perform effectively. Surfaces with a thickness of 7 mm or more ensure strong magnetic attraction and stability, making them ideal for optimal performance. On thinner surfaces (3 mm), the reduced magnetic attraction affected both stability and adherence, which could be critical for ensuring reliable operation in environments with varying surface conditions. This experiment suggests that surface thickness should be a consideration during deployment planning, especially in industries where thinner steel surfaces are common.

*Significance*: These findings emphasize the importance of surface characteristics in the stability of robot performance. For applications where surface thickness may vary, adjustments to the magnetic adhesion system or operational parameters may be required to maintain performance. Figure 13 summarizes the robot's performance across the different surface thicknesses, demonstrating the relationship between surface thickness and magnetic adhesion.

C. *Experiment 3: Incline Performance*

*Objective and Methodology*: The objective of this experiment was to evaluate the robot's performance on inclined surfaces by assessing its ability to maintain speed, stability, and magnetic adhesion at various incline angles. This experiment is crucial for determining the robot's effectiveness in environments where steel surfaces are not flat, and inclines of varying steepness may pose challenges for stability and movement.

*Procedure and Setup*: The robot was tested on a range of inclined surfaces, with angles starting at 15° and increasing incrementally to 45°. For each incline angle, the robot's speed, stability, and adhesion were recorded. These trials were conducted under uniform conditions to isolate the effect of the incline on performance. Speed was measured in meters per second, and stability was assessed based on whether the robot experienced slippage or struggled to maintain proper adhesion.

*Results and Observations*: The robot maintained full functionality on inclines up to 45°, though its performance varied with the angle. On smaller inclines (up to 15°), the robot exhibited minimal speed reduction and remained highly stable. As the incline increased beyond 30°, a gradual decrease in speed was observed, with the robot requiring more time to navigate the surface. At a 45° incline, although the robot remained operational, there was a noticeable reduction in both speed and stability, as the magnetic adhesion system had to work harder to maintain contact with the surface. Despite these challenges, the robot did not lose adhesion or experience significant slippage, demonstrating its capability to function effectively even on steep surfaces.

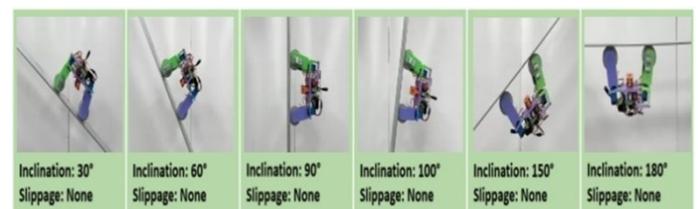

Fig. 14  Incline Performance

*Conclusion*: The experiment revealed that the robot is capable of operating on inclined surfaces up to 45°, although there is a trade-off between incline angle and operational efficiency. At steeper angles, the robot experiences a reduction in speed and requires more effort to maintain stability. These findings highlight the robustness of the magnetic adhesion system but also indicate that performance optimizations could be explored for steeper inclines to maintain efficiency.

*Significance*: Incline performance is critical for the robot's deployment in industrial environments where surfaces are rarely flat. Understanding the limits of its incline capabilities enables better planning and adaptation for various applications. These results underscore the importance of balancing speed and stability as incline angles increase, and they provide guidance for further refining the design for high-performance in steep environments. Figure 14 illustrates the relationship between incline angle and the robot's operational performance.

### D. Experiment 4: Omni-Directional Maneuverability

*Objective and Methodology*: The primary objective of this experiment was to assess the robot's ability to navigate complex environments requiring sharp turns and multi-directional movement. Given that the inspection robot will often encounter dynamic and unpredictable environments, ensuring accurate and responsive maneuverability in all directions is essential for effective inspection tasks. The experiment focused on evaluating the robot's precision in navigation, particularly during sharp turns and on curved paths.

*Procedure and Setup*: The robot was tested on a predefined track that included sharp 90-degree turns, S-shaped paths, and multi-directional changes. The robot's movements were tracked, and its ability to maintain accuracy without deviating from the planned path was recorded. The data collection process focused on deviations in trajectory, response time during directional changes, and the overall smoothness of movements.

*Results and Observations*: The robot demonstrated high accuracy in navigating complex paths, including sharp turns and curved trajectories, with minimal deviation. On 90-degree turns, the robot performed smoothly, maintaining precise control over its movements without any significant slippage or misalignment. The S-shaped paths introduced slight delays in movement due to the increased complexity of direction changes; however, these delays were minimal and within acceptable performance limits. The robot's ability to change directions without losing stability or accuracy was a clear indication of the efficiency of its omni-directional maneuvering capabilities.

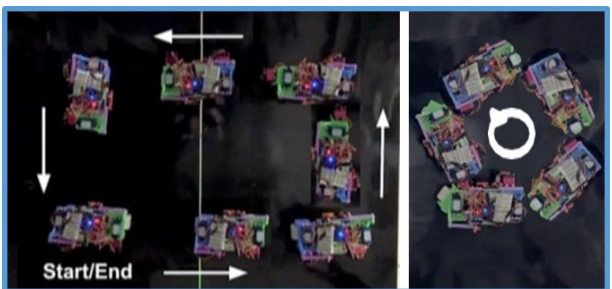

Fig. 15 Mobility Test

*Conclusion*: The robot exhibited advanced maneuverability, showcasing its ability to navigate intricate environments with high precision. While minor delays were observed in more complex paths, such as S-shaped turns, these delays did not significantly affect overall performance. The robot's omni-directional capabilities allow it to maintain accurate movement in challenging inspection environments, reinforcing its suitability for real-world applications where rapid changes in direction are required.

*Significance*: Omni-directional maneuverability is a critical feature for robots operating in environments where sharp turns and multi-directional movements are common, such as industrial plants or complex steel structures. The results from this experiment validate the robot's ability to handle such conditions with minimal deviation, providing confidence in its deployment for inspection tasks. Figure 15 illustrates the robot's performance across various directional paths, emphasizing its precision and adaptability in maneuvering through complex environments.

### E. Experiment 5: Speed Test

*Objective and Methodology*: The purpose of this experiment was to evaluate the robot's top speed on a flat surface under varying load conditions and battery levels. Understanding the robot's speed capability is essential for optimizing its operational efficiency in real-world scenarios. This test assessed how the robot's performance is influenced by changes in both load and battery power, providing critical data for ensuring that the system can maintain effective operation across a range of conditions.

*Procedure and Setup*: The robot was tested at different load levels, ranging from no load to the maximum load it could carry (27.5 kg), and at varying battery charge levels (100%, 75%, 50%, and 25%). Each combination of load and battery level was tested on a flat steel surface, and the robot's speed was measured in meters per second. Data was collected using high-precision timing equipment to ensure accuracy in measuring the speed across each scenario.

*Results and Observations*: The robot achieved its highest speed of 0.55 meters per second when carrying a light load and operating with a fully charged battery (100%). As the load increased, the speed gradually decreased due to the additional effort required to carry heavier weights. Similarly, reductions in battery levels also resulted in slower speeds, indicating the impact of lower power availability on motor performance. At the maximum load of 27.5 kg and a battery level of 25%, the robot's speed decreased significantly, demonstrating the combined effect of high load and low battery power on its performance.

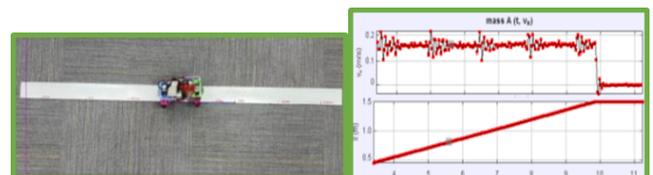

Fig. 16 Speed Test

*Conclusion*: The experiment revealed a clear relationship between the robot's speed, load, and battery level. While the robot can achieve a top speed of 0.55 m/s under optimal conditions (light load and full battery), its performance declines with heavier loads and lower battery levels. These findings are crucial for planning operational workflows in environments where the robot may need to carry equipment or operate for extended periods without recharging.

*Significance*: Understanding the robot's speed under various conditions allows for more efficient scheduling and deployment in real-world applications. The data from this experiment informs decisions about load management and battery optimization to ensure that the robot can maintain acceptable speeds during inspection tasks. Figure 16 illustrates the robot's speed performance across different load and battery conditions, providing a clear visual representation of how these factors influence overall operational efficiency.

*F. Experiment 6: Multi-Terrain Navigation*

*Objective and Methodology*: The goal of this experiment was to assess the robot's ability to navigate across various terrain types, including smooth metal surfaces, rusted areas, and debris-covered sections. Given that industrial inspection environments often feature diverse and unpredictable surface conditions, it is essential to verify the robot's adaptability and performance under these different terrains. The test aimed to evaluate the robot's speed, stability, and ability to handle obstacles.

*Procedure and Setup*: The robot was subjected to a series of navigation tests across different terrain types, including smooth metal surfaces, rusted steel, and areas with debris. Performance data, such as speed, stability, and obstacle navigation capability, were recorded for each terrain type. The terrain conditions were simulated to reflect real-world scenarios the robot may encounter during inspections. Data collection focused on measuring deviations in speed, slippage, and the robot's ability to clear obstacles without operator intervention.

*Results and Observations*: The robot demonstrated strong performance across all terrain types, maintaining stable adhesion and navigation capabilities on smooth metal surfaces. On rusted surfaces, the robot exhibited minor reductions in speed and stability due to the uneven and less magnetic surface texture. Similarly, debris-covered areas posed challenges by increasing friction and requiring the robot to navigate over small obstacles. Despite these challenges, the robot successfully navigated the debris-covered areas, overcoming obstacles with minimal deviation in its path and completing the inspection tasks without significant delays.

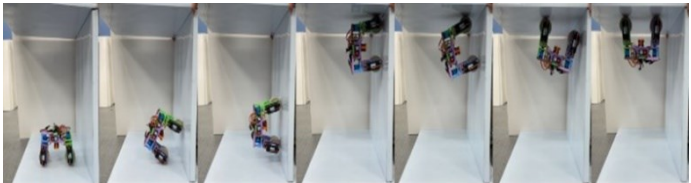

Fig 17. Navigation

*Conclusion:* The results indicate that the robot is highly adaptable and can successfully navigate diverse terrain conditions typically encountered in industrial inspection environments. While minor performance reductions were observed on rusted and debris-covered surfaces, the robot maintained functionality and was able to complete its tasks without significant loss of speed or stability. These results are significant for real-world deployments where varying surface conditions can affect performance.

*Significance*: The robot's ability to perform across different terrains demonstrates its versatility and robustness in complex environments. This adaptability is crucial for applications where surface conditions are unpredictable, such as in outdoor or industrial settings with corroded metal and debris. The ability to maintain stability and navigation in these environments ensures reliable performance and increases the operational range of the robot. Figure 17 illustrates the robot's performance across different terrain types, highlighting the minor reductions in speed and stability on rusted and debris-covered surfaces.

## VI. CONCLUSION

The development of the AI-Powered Magnetic Inspection Robot marks a significant advancement in the field of structural health monitoring (SHM), particularly for ferromagnetic structures such as bridges, pipelines, and storage tanks. This project has successfully integrated various technological components, enabling the robot to navigate a wide range of complex geometries, including flat surfaces, internal and external corners, cylindrical structures, and steep vertical inclines. The robust locomotion system, equipped with powerful magnetic adhesion and precision control mechanisms, ensures the robot maintains stability and reliability, even in severe or demanding environmental conditions.

One of the key successes of this project lies in its ability to solve the long-standing challenge of inspecting hard-to-reach areas of large-scale steel structures. Traditional inspection methods often prove inefficient, impractical, or cost-prohibitive in such scenarios. By leveraging robotic automation, this system not only reduces manual labor but also improves the accuracy and repeatability of defect detection, leading to more reliable maintenance and repair decisions. The machine learning models, particularly CNN and MobileNetV2, have demonstrated considerable success in identifying structural defects with precision, further enhancing the inspection process.

*Future Prospects*: Looking forward, several key areas of enhancement have been identified to expand the robot's potential:

- *Enhanced Mobility and Adaptability*: Future iterations could focus on improving the robot's locomotion capabilities, allowing for better performance across a broader range of environments, including more extreme terrains and surface conditions.
- *Real-time and Onboard Data Processing*: By integrating more advanced onboard processing, the robot could analyze data in real-time, allowing for quicker responses and more autonomous decision- making.
- *Expanded Defect Detection Capabilities*: With the integration of more sophisticated AI models and sensor systems, the robot's ability to detect a wider variety of defects could be greatly improved.
- *Autonomous Operation and Navigation*: Advancing the robot's autonomy will allow it to navigate and operate independently, minimizing human intervention and optimizing efficiency in large-scale industrial applications.

The experimental results from this project provide a strong foundation for future advancements in robotic inspection systems for SHM. These robots are poised to revolutionize the inspection processes, reducing costs, enhancing safety, and improving accuracy in maintaining critical infrastructure. With continued development, this AI-powered robot has the potential to open new frontiers, offering innovative solutions not only in traditional industries but also in emerging fields where inspection challenges are increasingly complex.

In conclusion, this project has set the stage for further advancements, focusing on continuous learning frameworks and on-device adaptation. This will ensure that the robotic system remains effective across a broad range of inspection scenarios, securing its role in the future of automated infrastructure inspection and maintenance. The possibilities for applying this technology across industries are vast, extending beyond current conceptions and offering exciting new prospects for the future.